\documentclass[11pt]{article}
\usepackage{acl2015}
\usepackage{amsmath}
\usepackage{amsfonts}
\usepackage{times}
\usepackage{url}
\usepackage{latexsym}
\usepackage{tikz}
\usepackage{graphicx}
\usepackage{subfigure}
\usepackage{multirow}
\usepackage{colortbl}
\usepackage{titlesec}
\usepackage{sistyle}
\SIthousandsep{,}

\usepackage{pgfplots}
\pgfplotsset{compat=1.3}

\definecolor{yellow2}{RGB}{166,97,26}
\definecolor{yellow1}{RGB}{223,194,125}
\definecolor{orange1}{RGB}{239,138,98}

\definecolor{gray_bg}{RGB}{230,230,230}
\definecolor{gray2}{RGB}{240,240,240}
\definecolor{gray1}{RGB}{77,77,77}
\definecolor{pink3}{RGB}{178,24,43}
\definecolor{pink2}{RGB}{239,138,98}
\definecolor{pink1}{RGB}{102,194,164}
\definecolor{white1}{RGB}{245,245,245}
\definecolor{green4}{RGB}{0,109,44}
\definecolor{green3}{RGB}{209,229,240}
\definecolor{green2}{RGB}{103,169,207}
\definecolor{green1}{RGB}{33,102,172}
\definecolor{dark_cyan}{rgb}{0.0, 0.55, 0.55}
\definecolor{medium_aquamarine}{rgb}{0.40,0.76,0.65}
\definecolor{cadmium_green}{rgb}{0.00,0.43,0.25}
\definecolor{matisse}{rgb}{0.13,0.45,0.68}
\definecolor{cardinal}{rgb}{0.75,0.15,0.21}
\definecolor{pale_chestnut}{rgb}{0.86,0.67,0.70}
\definecolor{pumice}{rgb}{0.78,0.78,0.78}
\definecolor{pastel_red}{rgb}{1, 0.4, 0.38}
\definecolor{biscay}{rgb}{0.07, 0.21, 0.42}

\title{Semantic Regularities in Document Representations}

\author{Fei Sun, Jiafeng Guo, Yanyan Lan, Jun Xu, \and Xueqi Cheng \\
  CAS Key Lab of Network Data Science and Technology\\
  Institute of Computing Technology\\
  Chinese Academy of Sciences, China \\
  {\tt ofey.sunfei@gmail.com}\\
  {\tt \{guojiafeng,\hspace{-2mm} lanyanyan,\hspace{-2mm} junxu,\hspace{-2mm} cxq\}@ict.ac.cn}
\\}

\date{}

\begin{document}
\maketitle

\begin{abstract}

Recent work exhibited that distributed word representations are good at capturing linguistic regularities in language.
This allows vector-oriented reasoning based on simple linear algebra between words.
Since many different methods have been proposed for learning document representations, it is natural to ask whether there is also linear structure in these learned representations to allow similar reasoning at document level.
To answer this question, we design a new document analogy task for testing the semantic regularities in document representations, and conduct empirical evaluations over several state-of-the-art document representation models.
The results reveal that neural embedding based document representations work better on this analogy task than conventional methods, and we provide some preliminary explanations over these observations.

\end{abstract}

\section{Introduction}

Recently, \newcite{Tomas:Linguistic} discovered that word representations learned by a recursive neural net (RNN) as well as by related log-linear models \cite{Mikolov:Distributed} can capture the linguistic regularities in language, which allows easy solutions to analogy questions of the form ``Beijing:China as Paris:\underline{\hspace{4mm}}'' using simple linear algebra.
With this word analogy task, a flurry of subsequent work exhibited that similar linear structure can also be revealed from representations learned from other methods \cite{Mnih:Learning,Jeffrey:GloVe,Levy:Linguistic}.

Besides word representation, document representation is also a fundamental and critical problem in natural language processing.
Over the past decades, various methods have been proposed to represent the document as a vector, including Bag of Words (BOW)~\cite{harris:Distributional}, Latent Semantic Indexing (LSI)~\cite{Deerwester:Indexing}, Non-negative Matrix Factorization (NMF)~\cite{Daniel:Learning}) and Latent Dirichlet Allocation (LDA)~\cite{Blei:Latent}.
Recently, there is a rising enthusiasm for applying the neural embedding methods to representing the documents~\cite{Nitish:Modeling,Quoc:Distributed}.

It is, therefore, natural to ask whether there is also linear structure in these learned document representations to allow similar reasoning at document level.
For example, given three articles talking about naive bayes, logistic regression, and hidden markov model, is it possible to find the article about conditional random fields as the solution to the document analogy question ``naive bayes: logistic regression as hidden Markov model: \underline{\hspace{4mm}}'' (\textit{i.e.}, document pairs explaining generative-discriminative model relations)?
Obviously, such reasoning is much more complex in semantics and cannot be achieved by simple retrieval or classification based on lexical information.
Representation with such linear structure would be useful for many semantic processing applications, \textit{e.g.}, it may help controversial search~\cite{Shiri:Navigating} by discovering document pairs talking about opposite facts on controversial topics with some seed pairs, or help non-local corpus navigation and paper recommendation together with word vectors~\cite{Dai:Document}.

For this purpose, we introduce a new document analogy task for evaluating the semantic regularities in document representations.
Since it is non-trivial to directly label the analogy questions over documents, we leverage the existing word/phrase semantic analogy test set and map the words/phrases in these questions to Wikipedia articles through title matching.
In this way, we obtain a large labeled analogy test set over documents.
The task is then to test whether different document representations over the Wikipedia articles can find the right answers to these semantic analogy questions.

Based on this test set, we evaluate several existing state-of-the-art document representations and show that neural embedding based models can achieve better performance than conventional models.
The major contributions of this paper include: 1) the introduction of a new document analogy task with benchmark dataset for evaluating document representations; 2) empirical comparison among state-of-the-art models and preliminary explanations over the results. 

\section{Measuring Semantic Regularities}
\begin{table*}[btp]
  \centering
  \renewcommand{\arraystretch}{0.917}
  \caption{Details of the test set, where the words/phases in example refers to Wikipedia articles.}
  \begin{tabular}{ l r l} \hline
    \textsc{Relation}  & \textsc{Count} & \textsc{Example} \\ \hline
    capital-common-countries & 506 & beijing: china $\sim$ paris: france \\
    capital-world & 3991 & bangkok: thailand $\sim$ cario: egypt\\
    currency & 88 & europe: euro $\sim$ india: rupee\\
    city-in-state & 277 & houston: texas $\sim$ miami: florida\\
    family & 56 & boy: girl $\sim$ man:woman\\
    newspapers & 20 & chicago: chicago tribune $\sim$ houston: houston chronicle\\
    ice hockey & 462 & boston: boston bruins $\sim$ los angeles: los angeles kings\\
    basketball & 306 & chicago: chicago bulls $\sim$ dallas: dallas mavericks\\
    airlines & 306 & canada: air canada $\sim$ italy: alitalia\\
    people-companies & 100 & bill gates: microsoft $\sim$ larry page: google\\\hline
  \end{tabular}
  \label{tab:da}
\end{table*}

\subsection{A Document Analogy Test Set}
\label{sec:dataset}
We propose to create a document analogy test set so that we can quantitatively evaluate how well different document representations capture semantic regularities.
Following the idea of word analogy task, we try to build a test set of analogy questions of the form ``$a$ is to $b$ as $c$ is to \underline{\hspace{4mm}}'', where $a, b, c$ are the identities of the documents.
However, it is not trivial to directly label the relations between two arbitrary documents due to the diversity in topics.
Fortunately, we found that each Wikipedia page is a concise document describing one specific concept, and thus the relations between the documents can be explained by their corresponding concepts.
Therefore, we can convert the task of labeling between documents into that between concepts (which are of words or phrases), where we already have a large labeled data set from \newcite{Mikolov:Distributed}.

Based on the idea above, we build a document analogy test set using Wikipedia and existing word and phrase analogy test set.
Specifically, we adopt the publicly available April 2010 dump of Wikipedia\footnote{\url{http://nlp.stanford.edu/data/WestburyLab.wikicorp.201004.txt.bz2}}~\cite{Shaoul:Westbury}, which has been widely used in~\cite{Huang:Improving,Luong:Better,Neelakantan:Efficient}.
The corpus contains \num{3035070} articles and about $1$ billion tokens.
We then collect all the existing word and phrase analogy test sets and match the words/phrases in questions to Wikipedia page titles.
Note here we do not take syntactic analogy questions of words into consideration because the relations between documents are usually semantic.
By resolving the ambiguity in matching, we finally obtain \num{6112} analogy questions over Wikipedia documents.
Table~\ref{tab:da} shows the details of the test set.

\subsection{Analogy Reasoning}

In this work, we adopt the same vector offset method~\cite{Tomas:Linguistic} for analogy reasoning.
To answer the questions like ``\textit{a} is to \textit{b} as \textit{c} is to \underline{\hspace{4mm}}'', we try to find a document with vector $\vec{x}$, which is the closest to $\vec{b} - \vec{a} + \vec{c}$ according to the cosine similarity:
\begin{equation}
    \arg \max_{\substack{x\in D, x\neq a\\ x \neq b,\ x\neq c} } (\vec{b} + \vec{c} - \vec{a})\cdot \vec{x}
    \label{eq:obj1}
\end{equation}
where $\vec{b},\, \vec{c},\, \vec{a}$ and $\vec{x}$ are the normalized document vectors.
The question is judged as correctly answered only if $x$ is exactly the answer document in the evaluation set.
The evaluation metric for this task is the percentage of questions answered correctly.

\section{Models}

In this section, we briefly summarize the models used in this paper.
Before that, we first list the notations.

Let $D{=}\{d_1,\ldots,d_N\}$ denote a corpus of $N$ documents over the word vocabulary $V{=}\{w_1,\ldots,w_{|V|}\}$.
Let $\mathbf{X} \in \mathbb{R}^{N\times |V|}$ be a document-word matrix, where entry $x_{ij}$ in $\mathbf{X}$ denotes the weight of the $j$-th word $w_j$ in the $i$-th document $d_j$.

\textbf{Bag of Words (BOW)} model treats a document as a bag (multiset) of its words.
It represents a document $d_i$ as a vector $\vec{x}_i=(x_{i1},\cdots,x_{i|V|})$, where $x_{ij}$ denotes the weight of the $j$-th word $w_j$ in the $i$-th document $d_j$.
The most popular weighting scheme for $x_{ij}$ is TF-IDF~\cite{Jones:statistical}.
However, The BOW model suffers from the sparsity and curse of dimensionality due to treating individual word as distinct feature.

\textbf{Matrix Factorization} methods attempt to tackle the limitation of BOW model through learning a low-dimensional vector for document by factorizing the document-word matrix $\mathbf{X}$.

\newcite{Deerwester:Indexing} applied truncated Singular Value Decomposition (SVD) to document-word matrix, namely Latent Semantic Indexing (LSI).
LSI approximates $\mathbf{X}$ by setting all but the largest $k$ singular values in $\mathbf{\Sigma}$ to $0$ ($\mathbf{\Sigma}_{k}$), as
\begin{equation*}
    \mathbf{X} \simeq \mathbf{D}\mathbf{\Sigma}_k \mathbf{W}^{\mathrm{T}}
\end{equation*}
Hence one might think of the rows of $\mathbf{D}\mathbf{\Sigma}_k$ as representations for documents in the latent space.

An alternative way is factorizing $\mathbf{X}$ into two non-negative matrices~\cite{Daniel:Learning},
\begin{eqnarray*}
    \mathbf{X} \simeq \mathbf{D}\mathbf{W}^{\mathrm{T}}
\end{eqnarray*}
where the rows of $\mathbf{D}$ can be seen as the representations of documents.

Unlike LSI which may have negative entries, NMF has better interpretability with the non-negative constraint.

\textbf{Topic Models} are also very popular in document representation fields because of their good interpretability, generalization ability and extensibility.
The most representative work is the Latent Dirichlet Allocation (LDA) model introduced by \newcite{Blei:Latent}.
It represents the documents as distributions over latent topics, where each topic is characterized by a distribution over words.

\textbf{Neural Embedding} models have attracted much attention in text representations due to its breakthrough in statistical language model~\cite{Bengio:Neural}.

The Paragraph Vector models are first introduced in~\cite{Quoc:Distributed} for document representation.
The Distributed Memory Model of Paragraph Vectors (PV-DM) captures the representation of a document via inserting a document vector to the continuous bag-of-words (CBOW) model~\cite{Tomas:Efficient}.
A simpler model can be obtained by replacing the input word vector with document vector in Skip Gram (SG) model, which is called ``Distributed Bag of Words version of Paragraph Vector'' (PV-DBOW).
 
\textbf{Bag of Word Embeddings (BOWE)} model tries to represent the document as a linear combination of word vectors, where the word vectors $\mathbb{W}$ can be obtained by tools like \texttt{Word2Vec} or \texttt{GloVe}. The low-dimensional representations of documents in BOWE can be written as
\begin{equation*}
    \mathbf{D} = \mathbf{X} \mathbf{W}
\end{equation*}
where $\mathbf{X}$ denotes the BOW representation of documents.

\section{Experiments}
\begin{table*}[btp]
  \centering
  \renewcommand{\arraystretch}{1.2}
  \caption{Results on the document analogy task under dimension 100. Bold scores are the best.}
  \begin{tabular}{ l r r r r >{\columncolor{gray2}[.85\tabcolsep]}r >{\columncolor{gray2}}r r} \hline
    \rowcolor{white}
    \textsc{Relation}  & BOW & LSI & NMF & LDA & PV-DM & PV-DBOW & BOWE\\ \hline
    capital-common-countries & 0.0 & 23.12 & 9.29 & 23.72 & 60.87 & 54.15 & \textbf{83.0}\\
    capital-world & 0.8 & 9.97 & 5.06 &  9.15 & 43.62 & 42.65 & \textbf{67.53}\\
    currency & 0.0 & 0.0 & 0.0 & 0.0 & 4.55 & 3.41 & \textbf{14.77}\\
    city-in-state & 0.0 & 7.94 & 6.50 & 4.33 & 33.57 & 34.30 & \textbf{51.26}\\
    family & 19.64 & 5.36  & 1.79 & 14.29 & \textbf{21.43} & \textbf{21.43} & 19.64\\
    newspapers & 5.0 & 25.0 & 10.0 & 10.0 & 5.0 & \textbf{50.0} & 40.0\\
    ice hockey & 0.0 & 3.68 & 2.16 & 0.0 & 12.12 & 20.13 & \textbf{33.33}\\
    basketball & 0.0 & 4.25 & 1.63 & 0.0 & 10.13 & 14.71 & \textbf{38.56}\\
    airlines & 11.76 &  9.15 & 2.94 & 2.61 & 12.42 & 20.26 & \textbf{42.48}\\
    people-companies & 2.0 & 1.0 & 0.0 & 0.0 & 6.0 & \textbf{12.0} & 2.0\\\hline
    total  & 1.34 & 9.88 & 4.81 & 8.43 & 37.47 & 37.76 & \textbf{60.42}\\\hline
  \end{tabular}
  \label{tab:d100}
\end{table*}
In this section, we first describe our experimental settings including the corpus, hyper-parameter selections, and specifications for different document representation methods.
Then we compare these methods on document analogy task and discuss the results.

\subsection{Corpus and Preprocessing}

The corpus used to learn document representations in this experiment is the same Wikipedia April 2010 dump as described in section~\ref{sec:dataset}.
In preprocessing, we lowercase the corpus, remove pure digit words, non-English characters and the words occur less than 20 times.

\subsection{Experimental Settings}

The baseline methods used in this paper including BOW with TF-IDF weight, LSI, NMF, LDA, PV-DM, PV-DBOW, and BOWE.
For BOW, LSI and LDA, we use the popular python topic model library \texttt{gensim}\footnote{\url{http://radimrehurek.com/gensim}}.
For NMF, we choose the python machine learning library \texttt{scikit learn}\footnote{\url{http://scikit-learn.org}}.
We implement PV-DM and PV-DBOW models in C++ due to \newcite{Quoc:Distributed} have not released source codes of PV models.
For word embeddings in BOWE, we use CBOW in the \texttt{Word2Vec} tool\footnote{\url{https://code.google.com/p/Word2Vec/}}.
The negative sampling method is adopted to take the place of the hierarchal softmax since we found the former always achieves better performance.
The learning rate is linearly decayed to $0$ as described in~\cite{Tomas:Efficient}, where the initial learning rate of PV-DM and CBOW model is 0.05, and PV-DBOW is 0.025.
We set context window size as 10 and use 10 negative samples.

\subsection{Results}
In Table~\ref{tab:d100}, we compare the results of 100-dimensional document vectors from all the methods on different subtasks of document analogy.
As we can see, among all the methods, BOW is almost the worst.
This demonstrates the weakness of simple vector space model on capturing semantic regularities.

Neural embedding models such as PV-DM and PV-DBOW perform much better than conventional latent models such as LSI, NMF, and LDA.
This is quite amazing since PV models can also be viewed as implicit matrix factorization according to the explanations on \texttt{Word2Vec}~\cite{Omer:Neural}.
A major difference is that conventional latent models usually work on matrix with each entry standing for the frequency or TF-IDF of a word in a document, while PV models factorize a shifted pointwise mutual information (shifted-PMI) matrix.
As discussed in~\cite{Sanjeev:Random}, PMI is a key factor why \texttt{Word2Vec} can work well for word analogy task.
We guess this might also be a major factor to explain the gap between PV models and other latent models.
Therefore, we conducted further experiments on LSI.
As a result, we find that the total accuracy of LSI can achieve 15.53\% with PMI matrix (about 57\% performance gain over LSI with TF-IDF matrix).
The results indicate that PMI plays an important role in revealing linear structure in document representations.

A surprising result is that the simple BOWE model performs significantly better than any other methods on almost all the subtasks ($p$-value $< 0.01$).
There might be two possible reasons for the result.
Firstly, when learning word vectors alone with \texttt{Word2Vec}, one can achieve very high scores on word analogy tasks~\cite{Tomas:Linguistic}.
BOWE thus benefits from the strong linear structures in word vectors by directly using word vectors as the representation of a document.
Secondly, the calculation of Euclidean distance\footnote{Note that dot product between normalized vectors in analogy reasoning is equivalent to Euclidean distance.} between documents under BOWE is equivalent to using a relaxed Word Mover's Distance~\cite{Matt:From}, which has been shown strong performance in measuring document distance.

We also conduct the experiments on different dimensions as shown in Table~\ref{tab:total}.
Similar trending can be found as that in Table~\ref{tab:d100}.

\begin{table}
  \centering
  \renewcommand{\arraystretch}{1.2}
  \caption{Results on the document analogy task under different dimensions.}
  \begin{tabular}{l r r r r} \hline
     & \multicolumn{4}{c}{\textsc{Dimension}} \\ \cline{2-5}
    \multirow{-2}{*}{\textsc{Model}}   & 50 & 100 & 150 & 200 \\\hline
   BOW & 1.34 & 1.34 & 1.34 & 1.34 \\
   LSI & 4.19 & 9.88 & 17.0 & 21.81 \\
   NMF & 1.59 & 4.81 & 8.75 & 11.85 \\
   LDA & 3.52 & 8.43 & 10.39 & 10.45 \\
   \rowcolor{gray2}
   \cellcolor{white}PV-DM & 25.62 & 37.47 & 37.71 & 36.03\\
   \arrayrulecolor{white}\hline
   \rowcolor{gray2}
   \cellcolor{white}PV-DBOW & 25.33 & 37.76 & 40.61 & 39.09\\
   BOWE & \textbf{42.05} & \textbf{60.42} & \textbf{66.74} & \textbf{69.49} \\
   \arrayrulecolor{black}\hline
  \end{tabular}
  \label{tab:total}
\end{table}

\section{Conclusion}

In this paper, we introduce a new document analogy task for quantitatively evaluating how well different document representations capture semantic regularities.
Based on the introduced benchmark dataset, we conduct empirical comparisons among several state-of-the-art document representation methods.
The results reveal that neural embedding based document representations work better on this analogy task.
We provide some preliminary explanations on these observations, leaving the inherent differences of these models to be further investigated in the future.
With this benchmark dataset, it would also be easier for us to develop new document representation models and to compare with existing methods.




\begin{thebibliography}{}

\bibitem[\protect\citename{Arora \bgroup et al.\egroup }2015]{Sanjeev:Random}
Sanjeev Arora, Yuanzhi Li, Yingyu Liang, Tengyu Ma, and Andrej Risteski.
\newblock 2015.
\newblock Random walks on context spaces: Towards an explanation of the
  mysteries of semantic word embeddings.
\newblock {\em CoRR}, abs/1502.03520.

\bibitem[\protect\citename{Bengio \bgroup et al.\egroup }2003]{Bengio:Neural}
Yoshua Bengio, R{\'e}jean Ducharme, Pascal Vincent, and Christian Janvin.
\newblock 2003.
\newblock A neural probabilistic language model.
\newblock {\em J. Mach. Learn. Res.}, 3:1137--1155, March.

\bibitem[\protect\citename{Blei \bgroup et al.\egroup }2003]{Blei:Latent}
David~M. Blei, Andrew~Y. Ng, and Michael~I. Jordan.
\newblock 2003.
\newblock Latent dirichlet allocation.
\newblock {\em J. Mach. Learn. Res.}, 3:993--1022, March.

\bibitem[\protect\citename{Dai \bgroup et al.\egroup }2014]{Dai:Document}
Andrew~M Dai, Christopher Olah, Quoc~V Le, and Greg~S Corrado.
\newblock 2014.
\newblock Document embedding with paragraph vectors.
\newblock {\em NIPS Deep Learning Workshop}.

\bibitem[\protect\citename{Deerwester \bgroup et al.\egroup
  }1990]{Deerwester:Indexing}
Scott Deerwester, Susan~T. Dumais, George~W. Furnas, Thomas~K. Landauer, and
  Richard Harshman.
\newblock 1990.
\newblock Indexing by latent semantic analysis.
\newblock {\em Journal of the American Society for Information Science},
  41(6):391--407.

\bibitem[\protect\citename{Dori-Hacohen \bgroup et al.\egroup
  }2015]{Shiri:Navigating}
Shiri Dori-Hacohen, Elad Yom-Tov, and James Allan.
\newblock 2015.
\newblock Navigating controversy as a complex search task.
\newblock In {\em Proceedings of the First International Workshop on Supporting
  Complex Search Tasks co-located with the 37th European Conference on
  Information Retrieval(ECIR 2015)}. Elsevier, March.

\bibitem[\protect\citename{Harris}1954]{harris:Distributional}
Zellig Harris.
\newblock 1954.
\newblock Distributional structure.
\newblock {\em Word}, 10(23):146--162.

\bibitem[\protect\citename{Huang \bgroup et al.\egroup }2012]{Huang:Improving}
Eric~H. Huang, Richard Socher, Christopher~D. Manning, and Andrew~Y. Ng.
\newblock 2012.
\newblock Improving word representations via global context and multiple word
  prototypes.
\newblock In {\em Proceedings of the 50th Annual Meeting of the Association for
  Computational Linguistics: Long Papers - Volume 1}, ACL '12, pages 873--882,
  Stroudsburg, PA, USA. Association for Computational Linguistics.

\bibitem[\protect\citename{Jones}1972]{Jones:statistical}
Karen~Spärck Jones.
\newblock 1972.
\newblock A statistical interpretation of term specificity and its application
  in retrieval.
\newblock {\em Journal of Documentation}, 28(1):11--21.

\bibitem[\protect\citename{Kusner \bgroup et al.\egroup }2015]{Matt:From}
Matt~J. Kusner, Yu~Sun, Nicholas~I. Kolkin, and Kilian~Q. Weinberger.
\newblock 2015.
\newblock From word embeddings to document distances.
\newblock In {\em Proceedings of the 32th International Conference on Machine
  Learning (ICML-15)}, ICML '15. ACM, New York, NY, USA, July.

\bibitem[\protect\citename{Le and Mikolov}2014]{Quoc:Distributed}
Quoc Le and Tomas Mikolov.
\newblock 2014.
\newblock Distributed representations of sentences and documents.
\newblock In Tony Jebara and Eric~P. Xing, editors, {\em Proceedings of the
  31st International Conference on Machine Learning (ICML-14)}, pages
  1188--1196. JMLR Workshop and Conference Proceedings.

\bibitem[\protect\citename{Lee and Seung}1999]{Daniel:Learning}
Daniel~D. Lee and H.~Sebastian Seung.
\newblock 1999.
\newblock Learning the parts of objects by non-negative matrix factorization.
\newblock {\em Nature}, 401(6755):788--791, october.

\bibitem[\protect\citename{Levy and Goldberg}2014a]{Omer:Neural}
Omer Levy and Yoav Goldberg.
\newblock 2014a.
\newblock Neural word embedding as implicit matrix factorization.
\newblock In {\em Advances in Neural Information Processing Systems 27}, pages
  2177--2185. Curran Associates, Inc., Montreal, Quebec, Canada.

\bibitem[\protect\citename{Levy and Goldberg}2014b]{Levy:Linguistic}
Omer Levy and Yoav Goldberg, 2014b.
\newblock {\em Proceedings of the Eighteenth Conference on Computational
  Natural Language Learning}, chapter Linguistic Regularities in Sparse and
  Explicit Word Representations, pages 171--180.
\newblock Association for Computational Linguistics.

\bibitem[\protect\citename{Luong \bgroup et al.\egroup }2013]{Luong:Better}
Minh-Thang Luong, Richard Socher, and Christopher~D. Manning.
\newblock 2013.
\newblock Better word representations with recursive neural networks for
  morphology.
\newblock In {\em Proceedings of the Seventeenth Conference on Computational
  Natural Language Learning}, pages 104--113. Association for Computational
  Linguistics.

\bibitem[\protect\citename{Mikolov \bgroup et al.\egroup
  }2013a]{Tomas:Efficient}
Tomas Mikolov, Kai Chen, Greg Corrado, and Jeffrey Dean.
\newblock 2013a.
\newblock Efficient estimation of word representations in vector space.
\newblock In {\em Proceedings of Workshop of ICLR}.

\bibitem[\protect\citename{Mikolov \bgroup et al.\egroup
  }2013b]{Mikolov:Distributed}
Tomas Mikolov, Ilya Sutskever, Kai Chen, Greg~S Corrado, and Jeff Dean.
\newblock 2013b.
\newblock Distributed representations of words and phrases and their
  compositionality.
\newblock In C.J.C. Burges, L.~Bottou, M.~Welling, Z.~Ghahramani, and K.Q.
  Weinberger, editors, {\em Advances in Neural Information Processing Systems
  26}, pages 3111--3119. Curran Associates, Inc.

\bibitem[\protect\citename{Mikolov \bgroup et al.\egroup
  }2013c]{Tomas:Linguistic}
Tomas Mikolov, Wen tau Yih, and Geoffrey Zweig.
\newblock 2013c.
\newblock Linguistic regularities in continuous space word representations.
\newblock In {\em Proceedings of the 2013 Conference of the North American
  Chapter of the Association for Computational Linguistics: Human Language
  Technologies (NAACL-HLT-2013)}. Association for Computational Linguistics,
  May.

\bibitem[\protect\citename{Mnih and Kavukcuoglu}2013]{Mnih:Learning}
Andriy Mnih and Koray Kavukcuoglu.
\newblock 2013.
\newblock Learning word embeddings efficiently with noise-contrastive
  estimation.
\newblock In C.J.C. Burges, L.~Bottou, M.~Welling, Z.~Ghahramani, and K.Q.
  Weinberger, editors, {\em Advances in Neural Information Processing Systems
  26}, pages 2265--2273. Curran Associates, Inc.

\bibitem[\protect\citename{Neelakantan \bgroup et al.\egroup
  }2014]{Neelakantan:Efficient}
Arvind Neelakantan, Jeevan Shankar, Alexandre Passos, and Andrew McCallum.
\newblock 2014.
\newblock Efficient non-parametric estimation of multiple embeddings per word
  in vector space.
\newblock In {\em Proceedings of the 2014 Conference on Empirical Methods in
  Natural Language Processing (EMNLP)}, pages 1059--1069, Doha, Qatar, October.
  Association for Computational Linguistics.

\bibitem[\protect\citename{Pennington \bgroup et al.\egroup
  }2014]{Jeffrey:GloVe}
Jeffrey Pennington, Richard Socher, and Christopher~D. Manning.
\newblock 2014.
\newblock Glove: Global vectors for word representation.
\newblock In {\em Proceedings of the 2014 Conference on Empirical Methods in
  Natural Language Processing, {EMNLP} 2014, October 25-29, 2014, Doha, Qatar,
  {A} meeting of SIGDAT, a Special Interest Group of the {ACL}}, pages
  1532--1543.

\bibitem[\protect\citename{Shaoul and Westbury}2010]{Shaoul:Westbury}
Cyrus Shaoul and Chris Westbury.
\newblock 2010.
\newblock The westbury lab wikipedia corpus.
\newblock {\em Edmonton, AB: University of Alberta}.

\bibitem[\protect\citename{Srivastava \bgroup et al.\egroup
  }2013]{Nitish:Modeling}
Nitish Srivastava, Ruslan Salakhutdinov, and Geoffrey~E. Hinton.
\newblock 2013.
\newblock Modeling documents with deep boltzmann machines.
\newblock In {\em Proceedings of the Twenty-Ninth Conference on Uncertainty in
  Artificial Intelligence}, pages 616--625, Seattle, USA, August.

\end{thebibliography}
\end{document}